\documentclass[runningheads]{llncs}
\usepackage{soul}
\usepackage{url}
\usepackage[hidelinks]{hyperref}
\usepackage[utf8]{inputenc}
\usepackage[small]{caption}
\usepackage{graphicx}
\usepackage{amsmath}
\usepackage{booktabs}
\usepackage{algorithm}
\usepackage{algorithmic}
\usepackage{color}
\usepackage{amsfonts}
\usepackage{booktabs}
\usepackage{array}
\usepackage{subfigure}
\def \forgetupdateblock{forget-update block}

\def \forgetupdatemodule{forget-update module}
\def \Forgetupdatemodule{Forget-update module}
\def \forgetblock{forget block}
\def \Forgetblock{Forget block}
\def \updateblock{update block}

\def \cross{cross-domain scenario}
\def \filtersize{filter\_size}
\def \supportquery{support samples and a query sample}

\def \matchingnet{MatchingNet}
\def \protopyticalnet{ProtoNet}
\def \relationnet{RelationNet}
\def \snail{SNAIL}
\def \maml{MAML}

\def \cchdel#1{{\st{#1}}}

\def \cchhl#1{{\color{blue}\hl{#1}}}
\def \nth#1#2{$#1$-#2}
\def \channelsequence{channel vector sequence}
\def \channelvectorconstruct{channel vector sequence construction module}

\begin{document}
\title{Channel Relationship Prediction with Forget-Update Module for Few-shot Classification}

\author{Minglei Yuan \and
Cunhao Cai \and
Tong Lu}

\institute{ National Key Lab for Novel Software Technology, Nanjing University, China 
\email{mlyuan@smail.nju.edu.cn} \\
\email{snowyjuneyuni@gmail.com} \\
\email{lutong@nju.edu.cn}}
%
%

%
%
%
\maketitle              
\begin{abstract}
In this paper, we proposed a pipeline for inferring the relationship of each class in support set and a query sample using \forgetupdatemodule{}. We first propose a novel architectural module called "\channelvectorconstruct{}", which boosts the performance of sequence-prediction-model-based few-shot classification methods by collecting the overall information of all \supportquery{}. The channel vector sequence generated by this module is organized in a way that each time step of the sequence contains the information from the corresponding channel of all support samples and the query sample to be inferred. Channel vector sequence is obtained by a convolutional neural network and a fully connected network, and the spliced \channelsequence{} is spliced of the corresponding channel vectors of \supportquery{} in the original channel order. Also, we propose a \forgetupdatemodule{} consisting of stacked \forgetupdateblock{}s. The \forgetblock{} modify the original information with the learned weights and the \updateblock{} establishes a dense connection for the model. The proposed pipeline, which consists of \channelvectorconstruct{} and \forgetupdatemodule{}, can infer the relationship between the query sample and support samples in few-shot classification scenario. Experimental results show that the pipeline can achieve state-of-the-art results on {\it mini}Imagenet, CUB dataset, and \cross{}.

\keywords{Few-shot classification  \and Sequence prediction \and Forget-update module.}
\end{abstract}
\section{Introduction}

Deep learning models often encounter the problem of overfitting when the amount of training data is limited, for example, one can expect that the given training data is insufficient compared to the complexity of deep models. The idea of meta-learning \cite{metalearning-bengio1992} can be employed to alleviate this problem. Meta-learning methods randomly sample data from the training set to simulate a test scenario, which is called a task or an episode hereinafter. Note that one episode contains a support set and a query set, the former of which consisting of a small number of labeled samples, while the latter consisting of unlabeled samples with labels to be predicted. In a meta-learning paradigm, thousands of randomly constructed tasks are used to train the model so that learned parameters can extract transferable knowledge, which makes a learned model generalize on unseen tasks.

On the basis of meta-learning, many recently proposed few-shot classification methods \cite{maml,siamese,matching,protonet,relationnet,snail} have gained better generalization abilities than universal deep learning models in few-shot paradigms. Some of them \cite{matching,protonet,relationnet} make use of distance metrics to measure distances between embeddings of two samples; however, they can not make full use of the overall information of \supportquery. \cite{snail} treats few-shot classification as a sequence-to-sequence problem, which does not suffer from the shortcomings of those using distance metrics. Notably, \cite{snail} proposes an attentive meta-learner called SNAIL with the help of temporal convolutions \cite{tcn} and soft attention \cite{attentionisallyouneed}. It claims to be able to learn a more flexible strategy. However, when dealing with few-shot learning tasks, SNAIL requires the embeddings of samples to be fed into the meta-learner as a sequence with each time step of the sequence being a sample-label pair. However, samples in each task are not spread across time, therefore it is hard for SNAIL to learn a feasible model. It is also reported in \cite{snail} that the authors didn't train the model successfully with an LSTM-based \cite{lstm} meta-learner on {\it mini}Imagenet \cite{matching} with a 4-layer feature extractor. 

To overcome the problem that some metric-based few-shot learning methods can not make full use of the overall information of \supportquery. In this paper we propose to construct a \channelsequence{} which is combined with the whole information of \supportquery. In this paradigm, each sample is sent into a convolutional neural network, and a corresponding multi-channel feature map is generated. After that, each feature map is converted into a embedding vector (We will call this vector as \textbf{channel vector}, hereinafter) of a specific scale through a fully connected network. At last, the \channelsequence{} is spliced by the corresponding channel vectors of \supportquery{} in the original channel order. After the \channelsequence{} is generated, the rest to do is to extract the internal relationship among the channels and use such information to perform few-shot classification. A natural idea is to apply LSTM \cite{lstm} or GRU \cite{gru} on \channelsequence{} to infer the relationship. 

Experimental results show that the spliced \channelsequence{} with LSTM or GRU model can get similar results with the state-of-the-art methods in few-shot classification tasks on CUB \cite{cub} and \cross{} \cite{closerfewshot}. This is because \channelsequence{} contains distinguish features which implicit the relationship between query sample and support set samples, and the sequence relationship can also be established on \channelsequence{} through back propagation method \cite{backpropagation}.  However, both LSTM and GRU have a disadvantage. They need to use the output of previous step as the input of current step, so none of these methods support parallel training. 

\begin{figure}[htbp]
\centering
\subfigure[Features not processed by forget-update module]{
\begin{minipage}{\linewidth}
\centering
\includegraphics[width=7cm]{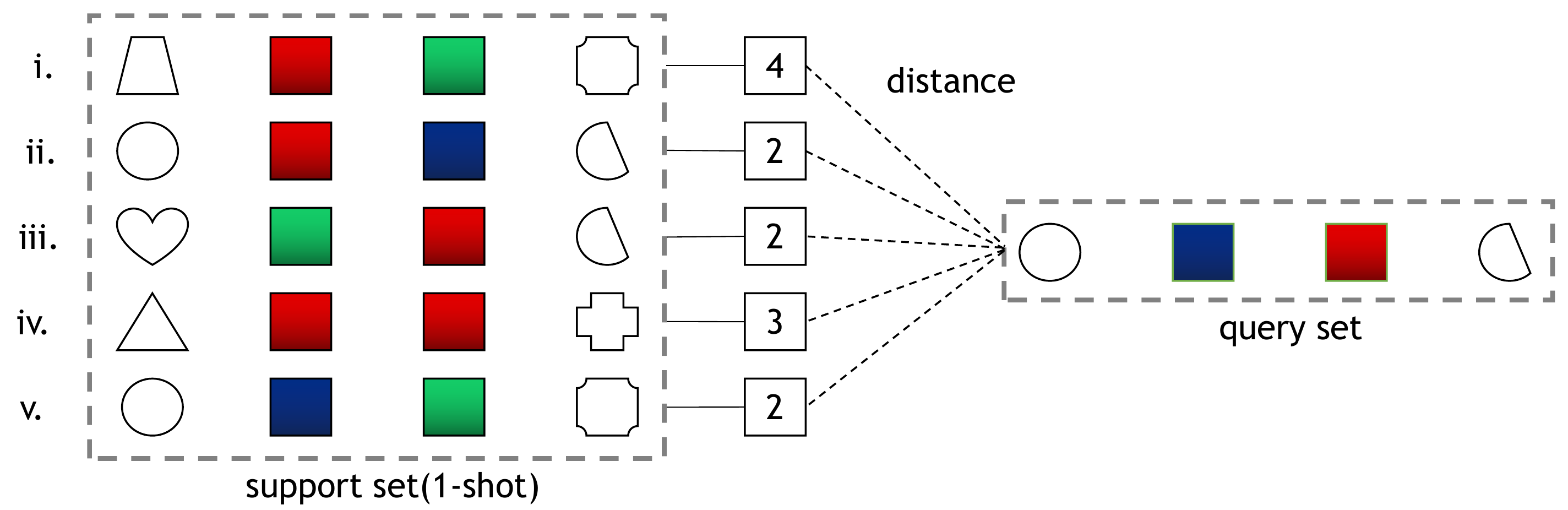}
\end{minipage}%
}%

\subfigure[Features processed by forget-update module]{
\begin{minipage}{\linewidth}
\centering
\includegraphics[width=7cm]{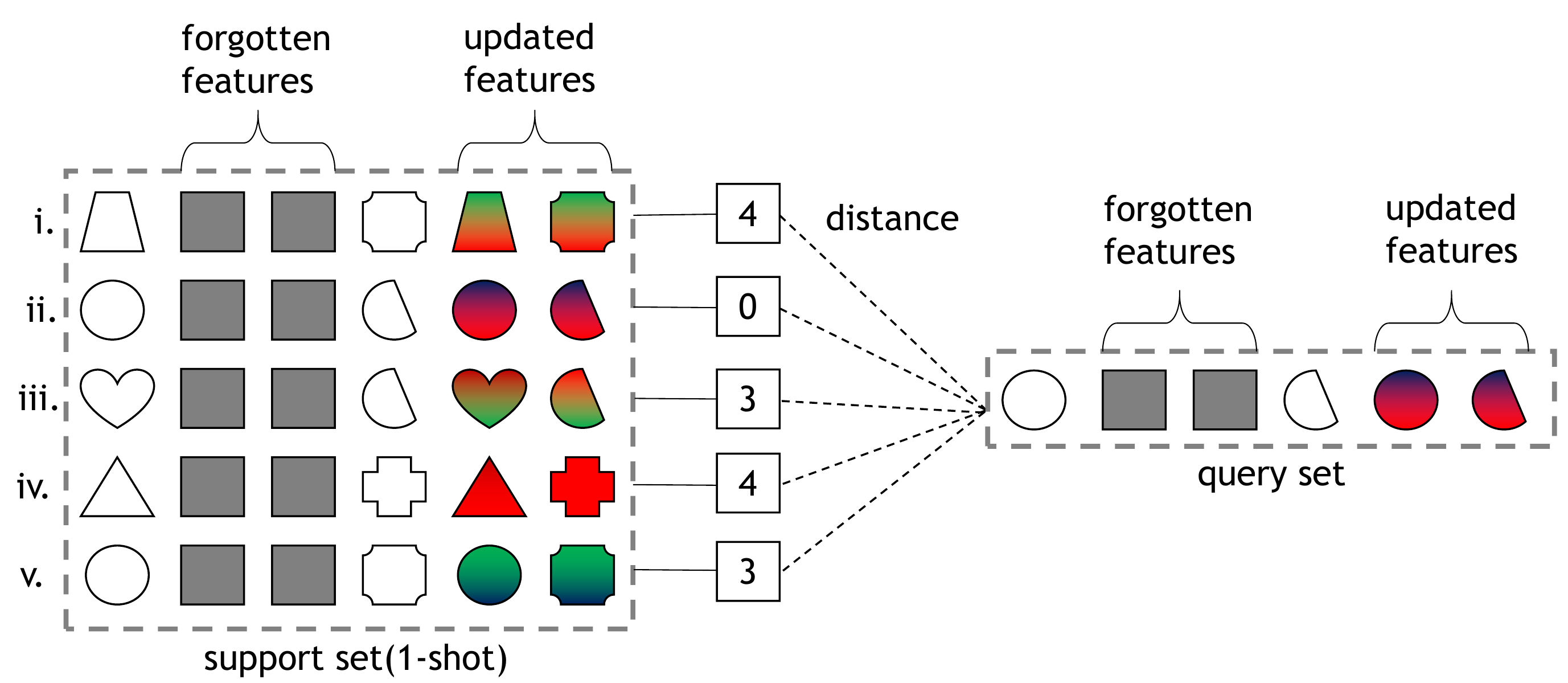}
\end{minipage}%
}%
\centering
\caption{A toy example illustrating the motivation of forget-update module. This is the 5-way 1-shot few-shot classification example.  The support set include five classes(i, ii, iii, iv, v), and the query set include one sample. (\textbf{a}) Each class include four features.  The distances between the query sample and each support class are 4, 2, 2, 3, 2. Under this condition, we cannot distinguish the category of the query sample. (\textbf{b}) Each class include six features, It includes forgotten features and updated features. The forgotten features are removed features that have low discrimination, while the updated features are newly generated features.}
\label{fig_motivation}
\end{figure}

To solve the problem that LSTM and GRU cannot train \channelsequence{} in parallel and get more distinguishing features, we proposed \forgetupdatemodule{} constructed by \forgetupdateblock{}s, which uses 1-dimensional causal dilated convolution as base block, so it can be trained in parallel. And every single \forgetupdateblock{} has two parts: a \forgetblock{} and a \updateblock{}. Fig. \ref{fig_motivation} give the motivation of \forgetupdatemodule{}. Specially, the \forgetblock{} equipped with a forget gated activation \cite{lstm} which can optimize existing information, and the \updateblock{} equipped with an update gated activation \cite{lstm} which can generate new information and establish dense connections. Experimental results demonstrated the effectiveness of \forgetupdatemodule{} on few-shot classification benchmarks. The proposed \forgetupdatemodule{} can get state-of-the-art results in few-shot classification tasks. 

The main contribution of this paper is twofold. First, we proposed to use the \channelsequence{} which contains the overall information of \supportquery{} to infer categories of query samples. When the proposed \channelsequence{} combined with sequence prediction methods, such as LSTM \cite{lstm} and GRU \cite{gru}, the relationship between the query samples and the support samples can be inferred. Experimental results show that we can get competitive results with the state-of-the-art few-shot learning methods on CUB \cite{cub} and \cross{} \cite{closerfewshot} when combines \channelsequence{} with LSTM \cite{lstm} or GRU \cite{gru}. Second, we design \forgetupdatemodule{} stacked of \forgetupdateblock{}s, which can produce more distinguish features in few-shot classification tasks. Experiments show that when combines the \forgetupdatemodule{} with \channelsequence{}, we can achieve state-of-the-art results on {\it mini}Imagenet \cite{matching}, CUB \cite{cub} and \cross{} \cite{closerfewshot}.

Code and datasets will be released once the article is accepted.

\section{Related Work}

\subsection{Metric-based Few-shot Leaning}
A number of few-shot classification methods are metric-based, i.e., they learn a set of projection functions that project the inputs to an embedding space, and a certain distance metric that measures the distance between any two embeddings. Those methods aim to make the samples from the same category closer in the embedding space, while those from different categories being distant from each other. For instance, Siamese Network \cite{siamese} extracts features from a pair of samples and calculates the similarity relationship between the two feature vectors, thus the classification is done by comparing the samples in the query set and the support set in such a manner. Prototypical Network \cite{protonet} is based on Euclidean distance metrics and uses the mean of embeddings from the same category as the prototype of that category. Relation Network \cite{relationnet} is similar to Prototypical Network, except that it employs a neural network to learn a deep instance metric, instead of using a fixed one.

\subsection{Meta-learner based Few-shot learning}
Some other methods construct a meta-learner that learn to make updates to the parameters to a traditional learner designed for scenarios with a great amount of data. \cite{maml} provided a method to initialize the parameters of the traditional learner in a way that a few gradient descent steps with a small amount of training data from a new task will lead to good generalization performance on that task. \cite{qi2018low} proposed to use the embedding vectors of the newly seen samples to imprint weights for the new classes on the rear of the base network. The traditional learner used in \cite{qiao2018few} is a convolutional-network-based network, and the method proposed in \cite{qiao2018few} learns to update the parameters of the last full-connected layer of the base network, based on the newly seen samples.


\subsection{Time Series Prediction Methods}
Aside from the recurrent neural networks (RNNs for short), a lot of recently proposed methods have been proven to work well in time series tasks, and some of them even show better performance than RNNs do. \cite{attentionisallyouneed} proposed a network architecture called Transformer to solve the sequence to sequence problem. The Transformer is built based on attention mechanism instead of RNNs, which makes it more parallelizable and requires less time to train. Recently, a 1-dimensional convolutional neural network architecture named TCN \cite{tcn} is proposed by Bai et al. and shows its effectiveness in sequence modeling problems. By combining causal convolution, dilated convolution, and shortcut connection, TCN is able to make use of information in a long history. 


\section{Proposed Approach}
\begin{figure}
    \centering
    \includegraphics[scale=0.46]{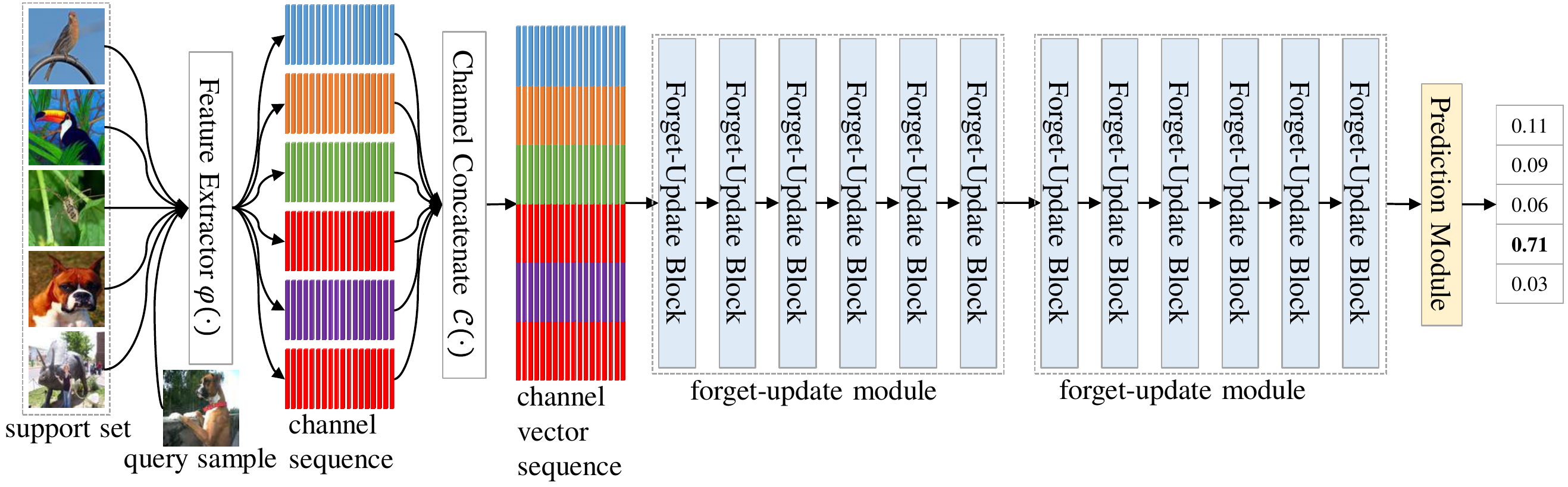}
    \caption{\textbf{The overall framework of the proposed method.} The feature extractor $\varphi(\cdot)$ first converts input images into $c$ matrices, and then converts each matrix into a 1-dimensional feature vector $m$, $m \in R^d$. The channel concatenate function $\mathcal{    C}(\cdot)$ is used to stitch all feature vectors and obtain \channelsequence{} $\widetilde{\mathbf{x}}$, $\widetilde{\mathbf{x}} \in R^{c\times ((N+1)\times d)}$, where $N$ is the class number of support set. Then, the \channelsequence{} $\widetilde{\mathbf{x}}$ is put into two \forgetupdatemodule{}s. Finally, the prediction module infers the similarity relationship between the query sample and each type of samples in the support set.}
     \label{fig_frame}
\end{figure}

In this section, after defining the problem, we describe the channel vector sequence construction module, the causal dilated convolution block, the forget-update module and the prediction module used in our method one by one in detail. The overall framework of the proposed method is shown in Fig.~\ref{fig_frame}. 

\subsection{Problem Definition}
\label{sec:prob_definition_3.1}

We first give the general setup and notations of few-shot classification in this paper. The purpose of few-shot classification is to build a model $\Phi(\cdot)$, which can classify unlabeled samples with the help of a few labeled samples. Each few-shot classification task $\mathcal{T}$(or episode) contains a labeled support set $\mathcal{S}$, an unlabeled query set $\mathcal{Q}$ and an output set $\mathcal{Y}$, which satisfies that the elements of $\mathcal{S}$ and $\mathcal{Q}$ do not intersect. The output set $\mathcal{Y}$ is the set of the labels of all the samples in $\mathcal{Q}$. We only consider $N$-way $K$-shot few-shot classification paradigm in this paper. In such a paradigm, every support set $\mathcal{S}$ contains exactly $N$ type of class and each class has $K$ samples, the query set $\mathcal{Q}$ contains some unlabeled samples that belong to the classes in $\mathcal{S}$. The output set $\mathcal{Y}$ includes the corresponding labels of elements in the query set $\mathcal{Q}$.

The support set $\mathcal{S}$, the query set $\mathcal{Q}$ and the output set $\mathcal{Y}$ are formalized as shown in Equation \ref{eq_support_set}, \ref{eq_query_set} and \ref{eq_out_put_set}.
\begin{equation}
	\label{eq_support_set}
	\begin{split}
	\mathcal{S} = \{(\mathbf{x}_{11},l_1),\cdots,(\mathbf{x}_{ij},l_i),\cdots,(\mathbf{x}_{NK},l_{N});\\
	l_i\in\{1,\cdots,N\} \}
    \end{split}
\end{equation}
\begin{equation}
	\label{eq_query_set}
	\mathcal{Q} = \{\mathbf{\dot{x}}_1,\cdots,\mathbf{\dot{x}}_{q}\}
\end{equation}
\begin{equation}
    \label{eq_out_put_set}
    \mathcal{Y} = (y_1,\cdots,y_{q}) \in \{1,\cdots,N\}^q
\end{equation}
where $N$ is the number of classes, $K$ is the sample number of each class in support set $\mathcal{S}$, and $q$ is the size of the query set. The subscripts of $\mathbf{x}_{ij}$ indicate that $\mathbf{x}_{ij}$ is the \nth{j}{th} sample in the \nth{i}{th} support class.

As in other state-of-the-art few-shot classification methods \cite{matching,maml,protonet,relationnet,snail}, our method trains a meta-learner to fit the few-shot classification task through minimizing the loss of its predictions over the query set $\mathcal{Q}$ as in Equation \ref{eq_optim}.
\begin{equation}
    \label{eq_optim}
    \theta^* = \underset{\theta}{\operatorname{argmin}} \mathbb{E}_{\mathcal{T}} \left[\sum_{\mathbf{\dot{x}}_i \in \mathcal{Q}, y_i \in \mathcal{Y}}\mathcal{L}(y_i, \Phi_{\theta}(\mathbf{\dot{x}}_i, \mathcal{S})) \right]
\end{equation}
where $\Phi(\cdot)$ indicates the meta-learner and $\mathcal{L}$ is the loss function. We need to train the meta-learner $\Phi(\cdot)$ with thousands of randomly sampled tasks under the constraint of loss function $\mathcal{L}(\cdot)$.

\subsection{Channel Vector Sequence Construction Module}
For each task $\mathcal{T}$, we first extract the feature map of each sample in the support set $\mathcal{S}$ and the query set $\mathcal{Q}$. When deal with $K$-shot tasks where $K>1$, for each class we calculate an element-wise average over the feature maps of all the samples in that class to form a class-level feature map. The class-level feature map for the \nth{i}{th} class can be formulated by Equation \ref{eq_average_embedding}.
\begin{equation}
    \label{eq_average_embedding}
    \mathbf{\bar{x}}_i^{'} = \frac{1}{N}\sum_{j=1}^{K}(\varphi(\mathbf{x}_{ij}))
\end{equation}
where $\varphi(\cdot) : \mathbb{R}^{c_{col}\times H\times W} \rightarrow \mathbb{R}^{c\times d}$ is a feature extractor ($c_{col}=1$ for grayscale images and $3$ for RGB images). We use $\varphi(\cdot) = \varphi_{sq}(\varphi_{conv}(\cdot))$ where $\varphi_{conv}(\cdot)$ is a 4-layer 2-dimensional convolution block and $\varphi_{sq}(\cdot)$ is a dimension reduction block. The output of $\varphi_{conv}(\cdot)$ is a feature map consisting of $c$ channels. $\varphi_{sq}(\cdot)$ then squeezes the information in each channel to a $d$-dimensional vector use two consecutive fully-connected layers. Note that we adopt $d=64$ and $c=64$ according to experiments.

After that, we perform channel-level stitching between all class-level feature maps in the support set $\mathcal{S}$ and feature maps of images in the query set $\mathcal{Q}$. Equation \ref{eq_channel_concat} shows how to stitch channels.

\begin{equation}
    \label{eq_channel_concat}
   \widetilde{\mathbf{x}}_{p} = \mathcal{    C}(\mathbf{\bar{x}}_1^{'},\cdots,\mathbf{\bar{x}}_N^{'},\varphi(\mathbf{\dot{x}}_{p}))
\end{equation}
where $\mathcal{    C}(\cdot)$ is the channel concatenate function which is used to splice $\mathbf{\bar{x}}_1^{'},\cdots,\mathbf{\bar{x}}_N^{'},\varphi(\mathbf{\dot{x}}_{p})$ according to the order of channels. $\widetilde{\mathbf{x}}_p \in \mathbb{R}^{c\times ((N+1)\times d)}$, $c$ is the number of channels, and $d$ is the feature dimension of each sample. $\widetilde{\mathbf{x}}_p$ is called \textbf{\channelsequence{}} in this paper.

Then, we transform few-shot classification problem into a sequence prediction problem on the \channelsequence{}. We formalize the prediction model as Equation \ref{eq_sequence_prediction}.
\begin{equation}
    \label{eq_sequence_prediction}
    \widetilde{y}_p = f(\widetilde{x}_{p1},\cdots,\widetilde{x}_{pc})
\end{equation}
where $p$ indicates \nth{p}{th} element in query set $\mathcal{Q}$, c is the \nth{c}{th} channel in $\widetilde{\mathbf{x}}_p$, $f(\cdot)$ is a sequence prediction model, $\widetilde{y}_p$ is the label of $\dot{\mathbf{x}}_p$.

\subsection{Causal Dilated Convolution Block}
\label{subsec_caual_dilated_conv}
The causal dilated convolution blocks are the basis of the proposed method. We use causal dilated convolution blocks on \channelsequence{}. Causal convolutions produce an output of the same length as the input and the newly generated data only depends on the data information before the current point. In addition, dilated convolution is also adopted to improve the range of receptive field on the \channelsequence{}. The dilation factor $d$ increased exponentially and can be formalized as Equation \ref{eq_dilation}. 

\begin{equation}
    \label{eq_dilation}
    d = k^{\ell-1}
\end{equation}
where $k$ is the kernel size and $\ell$ indicates the \nth{\ell}{th} layer of \forgetupdatemodule{}. The detail of \forgetupdatemodule{} is described in section \ref{sub_sec_update_forget_model}.

With the help of causal dilated convolution block, we can cover a very large receptive field with a few layers. Causal dilated convolution block is first used for generating raw audio in \cite{wavenet}.


\subsection{Forget-Update Module}
\label{sub_sec_update_forget_model}

\begin{figure}
    \centering
    \includegraphics[scale=0.65]{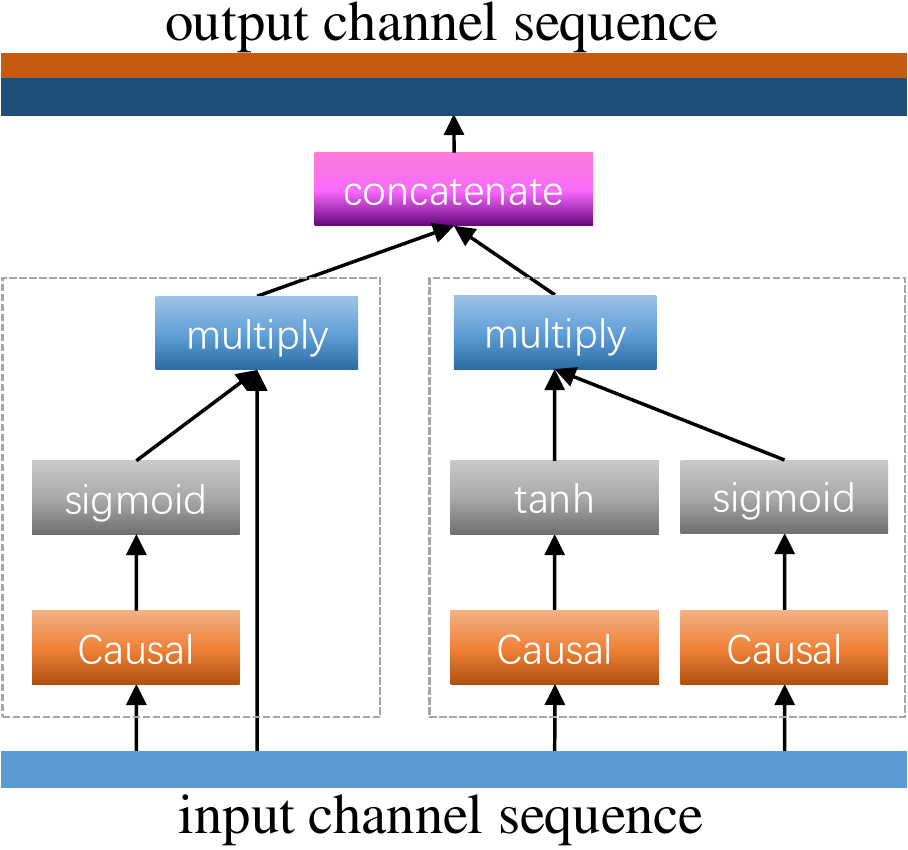}
    \caption{\textbf{Forget-update block.} The forget-update block includes two parts. The dashed boxes from left to right represent the forget block and update block, respectively. Causal indicates a causal dilated convolution block with dilated kernel size d = $k^{\ell-1}$.}
    \label{fig_forget_update}
\end{figure}

\Forgetupdatemodule{} consists of stacked \forgetupdateblock{}s. The detail of \forgetupdatemodule{} is described in Algorithm \ref{alg_memory_updata} and the \forgetupdateblock{} is illustrated in Fig.~\ref{fig_forget_update}. The \forgetblock{} is designed to upgrade existed information and the \updateblock{} is designed to reproduce new information and establish dense connection. The \forgetblock{} generates data $\mathbf{x}_{forget}(\mathbf{x}_{forget} \in \mathbb{R}^{c\times d_{in}})$ which has the same size as the input. \Forgetblock{} can be formalized as Equation \ref{eq_forget}.

\begin{equation}
    \label{eq_forget}
    \mathbf{x}_{forget} = \sigma(Causal(\widetilde{\mathbf{x}}^{(i)},d,k))\odot\widetilde{\mathbf{x}}^{(i)}
\end{equation}

where $Causal(\cdot)$ is causal dilated convolution function, $\sigma(\cdot)$ is a sigmoid function, $d$ is dilated rate, $k$ is kernel size, $\widetilde{\mathbf{x}}^{(i)}$ is the \textit{i}-th input to the \textit{i}-th \forgetupdateblock{} in \forgetupdatemodule{}.

The update model will generate data $\mathbf{x}_{update}(\mathbf{x}_{update}\in \mathbb{R}^{c\times {filter\_size}})$ which has the same sequence length as the input, and the dimension of each channel vector is set to $\filtersize$. The \channelsequence{} $\widetilde{\mathbf{x}}$ is used as the input of the first forget-update block in the proposed model. The output channel vector sequence will be used as input for the next \forgetupdateblock{}. The process of \updateblock{} can be formalized as Equation \ref{eq_update1},\ref{eq_update2},\ref{eq_update3}.
\begin{equation}
    \label{eq_update1}
    temp1 = \tanh({Causal(\widetilde{\mathbf{x}}^{(i)},d,k))})
\end{equation}

\begin{equation}
    \label{eq_update2}
    temp2 = \sigma(Causal(\widetilde{\mathbf{x}}^{(i)},d,k)))
\end{equation}

\begin{equation}
    \label{eq_update3}
    \mathbf{x}_{update} = temp1 \odot temp2
\end{equation}

where $tanh(\cdot)$ is hyperbolic tangent activation function.

And finally, we stitch $\mathbf{x}_{forget}$ and $\mathbf{x}_{update}$ in feature dimension direction as the output of \forgetupdateblock{}.

\begin{algorithm}[htb]
\caption{Forget-update module. $\sigma(\cdot)$ is the sigmoid function, $\odot$ denotes element-wise multiplication operator and $\mathcal{    C}(\cdot)$ is the channel concatenate function, $Causal(\cdot)$ is the causal dilated convolution function, which is detailed in Section \ref{sub_sec_update_forget_model}.}
\label{alg_memory_updata}
\begin{algorithmic}[1]
\REQUIRE ~~\\
$\widetilde{\mathbf{x}}$: \channelsequence{} \\
$k$: kernel size  \\
$c$: the length of channel sequence \\
\STATE $\widetilde{\mathbf{x}}^{(0)}$ = $\widetilde{\mathbf{x}}$
\STATE $\ell$ = $ \lceil\log_k c\rceil$
\STATE $i = 0$
\WHILE {$i < \ell$}
\STATE $d = k^{i}$
\STATE $\mathbf{x}_{forget} = \sigma(Causal(\widetilde{\mathbf{x}}^{(i)},d,k))\odot\widetilde{\mathbf{x}}^{(i)}$
\STATE $temp1 = \tanh({Causal(\widetilde{\mathbf{x}}^{(i)},d,k))})$
\STATE $temp2 = \sigma(Causal(\widetilde{\mathbf{x}}^{(i)},d,k)))$
\STATE $\mathbf{x}_{update} = temp1 \odot temp2$
\STATE $\widetilde{\mathbf{x}}^{(i+1)} = \mathcal{    C}(\mathbf{x}_{forget},\mathbf{x}_{update})$
\STATE $i = i + 1$
\ENDWHILE
\RETURN $\widetilde{\mathbf{x}}^{(\ell-1)}$
\end{algorithmic}
\end{algorithm}

\subsection{Prediction Module}
The prediction module is a three-layer fully connected network, which uses Relu \cite{relu} as the activation function. The weights of the prediction model are initialized using the method described in \cite{kaimingInitial} and weight normalization \cite{weightnormal} is applied. The prediction module predicts $N$ values which represent the similarity relationship between the query sample's feature map and the class-level feature maps of those $N$ support classes.

\section{Experiments and Discussion}

\begin{table}[ht]
\renewcommand\arraystretch{1.1}
\centering
\caption{5-way 1-shot and 5-way 5-shot classification accuracies on {\it mini}Imagenet. All results are averaged over 600 episodes from the test set. The top two results are highlighted. The unit is percentage. $^*$ indicates performing 30-way training in 1-shot tasks and 20-way training in 5-shot tasks during the training stage. The unit is percentage.}

\small

\begin{tabular}[t]{r|>{\centering}p{0.18\linewidth}>{\centering\arraybackslash}p{0.18\linewidth}}
\toprule
Methods     &1-shot&5-shot \\
\midrule
\matchingnet \cite{matching}                   & 46.6    & 60.0  \\
\maml \cite{maml}                          & 48.70   & 63.11 \\
\protopyticalnet$^{*}$          & 49.42   & \textbf{68.20} \\
\relationnet \cite{relationnet}                    & \textbf{50.44} & 65.32 \\
\snail{} \cite{snail}                        & 45.10 & 55.20     \\
\midrule
proposed                        & \textbf{50.54} & \textbf{65.66} \\
\bottomrule
\end{tabular}
\normalsize
\label{tab_mini}
\end{table}

\begin{table}[ht]
\renewcommand\arraystretch{1.1}
\centering
\caption{5-way 1-shot, 5-way 3-shot and 5-way 5-shot classification accuracies on CUB. All results are averaged over 600 episodes from the test set. The top two results are highlighted. The unit is percentage.}
\small
\begin{tabular}[t]{r|>{\centering}p{0.18\linewidth}>{\centering}p{0.18\linewidth}>{\centering\arraybackslash}p{0.18\linewidth}}
\toprule
Methods                    & 1-shot        & 3-shot        & 5-shot \\
\midrule
\matchingnet \cite{matching}       & 59.75    & 69.92    & \textbf{73.59} \\
\protopyticalnet \cite{protonet}    & 50.18    & 62.70	& 66.82 \\
\relationnet \cite{relationnet}        & \textbf{61.18}   & \textbf{70.40}   & 73.45 \\
\midrule
proposed            & \textbf{60.89}   & \textbf{73.03}   & \textbf{76.79} \\
\bottomrule
\end{tabular}
\normalsize
\label{tab_cub}
\end{table}

\begin{table}[ht]
\renewcommand\arraystretch{1.1}
\centering
\caption{5-way 1-shot, 3-shot and 5-shot classification accuracies on \cross. All results are averaged over 600 episodes from the test set. The top two results are highlighted. The unit is percentage.}
\small

\begin{tabular}[t]{r|>{\centering}p{0.18\linewidth}>{\centering}p{0.18\linewidth}>{\centering\arraybackslash}p{0.18\linewidth}}
\toprule
Methods    & 1-shot        & 3-shot        & 5-shot \\
\midrule
\matchingnet \cite{matching}    & 40.24    & 50.61    & 55.93 \\
\protopyticalnet \cite{protonet} & 37.24    & 50.19    & 56.49 \\
\relationnet \cite{relationnet}    & \textbf{41.41}    & \textbf{52.85}    & \textbf{58.93} \\
\midrule
proposed         & \textbf{46.48}    & \textbf{55.27}    & \textbf{57.69}  \\
\bottomrule
\end{tabular}
\normalsize
\label{tab_cross}
\end{table}

\begin{table*}[ht]
\renewcommand\arraystretch{1.1}
\centering
\caption{Ablation experiments on {\it mini}Imagenet, CUB and \cross. All results are averaged over 600 episodes from the test set. The top results are highlighted. The unit is percentage.}
\small
\begin{tabular}[t]{r|
                    >{\centering}p{0.1\linewidth}
                    >{\centering}p{0.1\linewidth}
                    >{\centering}p{0.1\linewidth}
                    >{\centering}p{0.1\linewidth}
                    >{\centering}p{0.1\linewidth}
                    >{\centering}p{0.1\linewidth}
                    >{\centering}p{0.1\linewidth}
                    >{\centering\arraybackslash}p{0.1\linewidth}}
\toprule
     Methods
     &\multicolumn{2}{c}{{\it mini}Imagenet} &\multicolumn{3}{|c}{CUB}  &\multicolumn{3}{|c}{cross} \\
     &\multicolumn{1}{c}{1-shot}
     &\multicolumn{1}{c}{5-shot}
     &\multicolumn{1}{|c}{1-shot}
     &\multicolumn{1}{c}{3-shot}
     &\multicolumn{1}{c}{5-shot}
     &\multicolumn{1}{|c}{1-shot}
     &\multicolumn{1}{c}{3-shot}
     &\multicolumn{1}{c}{5-shot} \\
\midrule
TCN \cite{tcn}   & 40.32    & 63.39
        & 57.76   & 58.07    & 60.47
        & 38.38   & 43.59    & 38.64  \\
+update  & 45.71    & 64.55
        & 54.05    & 70.19    & 71.54
        & 42.48    & 54.42    & \textbf{58.20} \\
proposed    & \textbf{50.54}   & \textbf{65.66}
            & \textbf{60.89}   & \textbf{73.03}   & \textbf{76.79}
            & \textbf{46.48}   & \textbf{55.27}   & 57.69 \\
\bottomrule
\end{tabular}
\normalsize
\label{tab_ablation}
\end{table*}


\begin{table*}[ht]
\renewcommand\arraystretch{1.1}
\centering
\caption{The prediction results of LSTM and GRU networks with the generated \channelsequence{} as input. All results are averaged over 600 episodes from the test set. The top two results are highlighted. The unit is percentage. $^*$ indicates performing 30-way training in 1-shot tasks and 20-way training in 5-shot tasks during the training stage. The unit is percentage.}
\small
\begin{tabular}[t]{r|
                    >{\centering}p{0.1\linewidth}
                    >{\centering}p{0.1\linewidth}
                    >{\centering}p{0.1\linewidth}
                    >{\centering}p{0.1\linewidth}
                    >{\centering}p{0.1\linewidth}
                    >{\centering}p{0.1\linewidth}
                    >{\centering}p{0.1\linewidth}
                    >{\centering\arraybackslash}p{0.1\linewidth}}
\toprule
    Methods
     &\multicolumn{2}{c|}{{\it mini}Imagenet} &\multicolumn{3}{c|}{CUB}  &\multicolumn{3}{c}{cross} \\
     &\multicolumn{1}{c}{1-shot}
     &\multicolumn{1}{c|}{5-shot}
     &\multicolumn{1}{c}{1-shot}
     &\multicolumn{1}{c}{3-shot}
     &\multicolumn{1}{c|}{5-shot}
     &\multicolumn{1}{c}{1-shot}
     &\multicolumn{1}{c}{3-shot}
     &\multicolumn{1}{c}{5-shot} \\
\midrule
\matchingnet    & 46.6              & 60.0
                & 59.75    & 69.92    & 73.59
                & 40.24    & 50.61    & 55.93 \\
\protopyticalnet
                & $49.42^*$   & {$\textbf{68.20}^*$}
                & 50.18    & 62.70    & 66.82
                & 37.24    & 50.19    & 56.49 \\
\relationnet    & \textbf{50.44}   & 65.32
                & \textbf{61.18}   & 70.40    & 73.45
                & 41.41            & 52.85    & 58.93 \\

\midrule

GRU             & 48.86    & 63.35
                & 59.07    & 71.57    & 74.29
                & 42.97    & 53.38    & \textbf{61.12} \\
                
LSTM            & 49.40            & 61.92
                & 60.28            & \textbf{71.69}    & \textbf{74.88}
                & \textbf{44.80}   & \textbf{54.74}    & \textbf{59.78} \\

\midrule
proposed        & \textbf{50.54}   & \textbf{65.66}
                & \textbf{60.89}   & \textbf{73.03}   & \textbf{76.79}
                & \textbf{46.48}   & \textbf{55.27}   & 57.69 \\
\bottomrule
\end{tabular}
\label{tab_sequence}
\end{table*}
\normalsize
In this section, we evaluate the proposed method and some state-of-the-art few-shot classification methods in three scenarios, including generic object recognition, fine-grained classification, and \cross{} classification.

\subsection{Datasets and Scenarios}
For the scenario of generic object recognition, we use the widely used {\it mini}Imagenet dataset \cite{matching}. {\it mini}Imagenet is a subset of Imagenet \cite{imagenet} and consists of 100 classes, each of which contains 600 images with a size of 84$\times$84. We follow the split used in \cite{relationnet} and split the dataset into 64, 16 and 20 classes for training, validation and testing, respectively.

For the scenario of fine-grained classification, we use the CUB dataset \cite{cub}. It contains 11,788 images from 200 classes in total. In our experiments, we split the dataset into 100, 50 and 50 classes for training, validation and testing, respectively.

For the \cross{} ({\it mini}Imagenet $\rightarrow$ CUB. For simplicity, we will call it \textbf{cross} hereinafter), we follow the setting used in \cite{closerfewshot}, which uses {\it mini}Imagenet as the training set while using the 50 CUB validation classes for validation and the 50 CUB testing classes for testing. This is to test out the performance of our method when the effect of domain shift is relatively significant.

\subsection{Implementation Details}

\cite{matching} first describes the meta-learning training setup in few-shot classification areas. All the methods in this paper are trained with meta-learning strategy. Specially, each prediction in a task (or episode) only relies on a corresponding support set. In the meta-training process, we train 60,000 episodes when using CUB or cross as the dataset, and 120,000 episodes with {\it mini}Imagenet. We adopt the $N$-way $K$-shot few-shot classification paradigm. In each episode, we randomly choose $N=5$ classes to use in this episode. We then randomly sample $K$ images for each previously chosen class to make up the support set. The size of the query set is fixed to 16. 1-shot and 5-shot classification are evaluated on {\it mini}Imagenet dataset. 1-shot, 3-shot and 5-shot classification are evaluated on CUB and cross datasets. In the meta-training process, the model, which has the best accuracy when evaluated in validation set, is saved. And the saved model is used to evaluate the test accuracy. In the meta-testing process, we test the model with 600 episodes and adopt the average of all the prediction results as the testing accuracy.


In this paper, all the methods are trained from scratch and adopt Adam \cite{adam} as optimizer. The optimizer takes an initial learning rate of 0.001, and we reduce the learning rate by 10\% when the testing accuracy stagnates in 7 consecutive training steps. For each convolutional layer in $\varphi_{conv}(\cdot)$, a Batchnorm \cite{batchnorm} regularization module is inserted between the convolution and the activation function. All the convolutions uses 64 $3\times 3$ kernels and the activation function is ReLU \cite{relu}. A $2\times 2$ max-pooling operation is added to the first two layers. We only apply normalization, scale and center-crop operation on the input images without data enhancement. We reimplemented \matchingnet{} \cite{matching}, \protopyticalnet{} \cite{protonet} and \relationnet{} \cite{relationnet} on CUB and cross. 

The proposed model contains two \forgetupdatemodule{}s. Every \forgetupdatemodule{} contains $\lceil\log_k c\rceil$ ($k=2,c=64$) \forgetupdateblock{}s. The \filtersize{} of all \forgetupdateblock{}s in the first and second \forgetupdatemodule{} is set to 16 and 32, respectively.

\subsection{Experiment Results}

Our Experiments are designed in the purpose of to answer these questions:
\begin{enumerate}
    \item How does the proposed method compare to exist methods?
    \item Evaluate the effects of forget block and update block.
    \item Evaluate the prediction results of LSTM \cite{lstm} and GRU \cite{gru} networks with \channelsequence{} as input.
\end{enumerate}
To validate the effectiveness of the proposed method, we compare it with {\textbf\matchingnet{}} \cite{matching}, \textbf{\protopyticalnet{}} \cite{protonet} and \textbf{\relationnet{}} \cite{relationnet} on {\it mini}Imagenet, CUB datasets, and \cross. In addition, we also compared with \textbf{\maml{}} \cite{maml} and \textbf{\snail{}} \cite{snail} on {\it mini}Imagenet.


Table \ref{tab_mini}, \ref{tab_cub}, \ref{tab_cross} illustrate the object recognition capability on miniImagenet, CUB and \cross{}, respectively. All the methods use a 4-layer convolutional network as backbone. Table \ref{tab_mini} shows that the proposed method achieves the best performance in 5-way 1-shot paradigm and achieves the second-best performance in 5-way 5-shot paradigm. \protopyticalnet{} has a big improvement over the proposed method. This is most likely because \protopyticalnet{} is trained in a 20-way 5-shot paradigm, and tested in a 5-way 5-shot paradigm, as a result, it gains greater discrimination on object recognition task in 5-way 5-shot paradigm. Table \ref{tab_cub} shows that the proposed method has made great progress on both 5-way 3-shot and 5-way 5-shot tasks and achieve a second place on 5-way 1-shot task. Table \ref{tab_cross} shows that the proposed method has a larger improvement on 5-way 1-shot and 5-way 3-shot tasks than the comparative methods, and achieves sub-optimal result on 5-way 5-shot task.

Table \ref{tab_ablation} shows the results of the ablation experiments. \textbf{TCN} \cite{tcn} is the baseline method which uses causal dilated convolution with identity connection \cite{resnet} and uses similar channel configuration as the proposed method. \textbf{+update} method uses the proposed update block but no forget block. The \textbf{proposed} method uses the \forgetupdatemodule{} and put \channelsequence{} as input. Table \ref{tab_ablation} shows that the forget block and update block can improve the prediction performance on {\it mini}Imagenet, CUB and cross.

Table \ref{tab_sequence} shows the prediction results of \textbf{GRU} \cite{gru} and \textbf{LSTM} \cite{lstm} methods with the proposed \channelsequence{} as input. LSTM and GRU hidden layer dimensions are both set to 512, the number of layers are set to 2 and bidirectional mode are both set to false. Table \ref{tab_sequence} shows that most results of LSTM and GRU methods have a better performance than the state-of-the-art few-show classification methods in CUB and \cross{}, and the performance of LSTM and GRU on {\it mini}Imagenet have competitive results with matchnet \cite{matching}. It reflects that the proposed \channelsequence{} can be combined with time sequence models, and this method can be used to infer the similarity relationship between query sample and support set samples. It also indicates that the proposed method almost completely exceeds LSTM and GRU, which means that the proposed \forgetupdatemodule{} is more suitable for the proposed \channelsequence{} than LSTM and GRU.


\subsection{Discussion}
From the experimental results, we find that the pipeline which uses \channelsequence{} as input and uses \forgetupdatemodule{} as relation discriminator can get state-of-the-art results for few-shot classification. We think that the \updateblock{} is equivalent to a dense connection mechanism, which can generate new information when passing through each \updateblock{}. The \forgetblock{} can modify existing information with learned weight. The combination of LSTM and GRU with the proposed \channelsequence{} can get state-of-the-art results on CUB and \cross{}, indicates that the similarity relationship between class-level features and query sample's feature is implicated in the proposed \channelsequence{}.

\section{Conclusion}
In this paper, we study the spliced \channelsequence{} of \supportquery. Experimental results show that putting the spliced \channelsequence{} as the input of LSTM and GRU can get competitive results with state-of-the-art few-shot classification methods in the CUB dataset and cross-domain scenario ({\it mini}Imagenet $\rightarrow$ CUB). This shows that the sequence prediction methods can be used to infer the similarity relationship with the help of \channelsequence{}. We also proposed a \forgetupdatemodule{}, and the proposed module shows promising results on few-shot classification tasks when using the proposed \channelsequence{} as input.

%
%
%

\end{document}